\documentclass[lettersize,journal]{IEEEtran}
\usepackage{amsmath,amsfonts}
\usepackage{algorithmic}
\usepackage{array}
\usepackage[caption=false,font=normalsize,labelfont=sf,textfont=sf]{subfig}
\usepackage{textcomp}
\usepackage{stfloats}
\usepackage{url}
\usepackage{verbatim}
\usepackage{graphicx}
\usepackage{multirow}

\usepackage{multicol}
\usepackage[table,xcdraw]{xcolor}
\usepackage{xcolor}
\definecolor{shadecolor}{rgb}{1,.8,.3}
\newcommand*{\Scale}[2][4]{\scalebox{#1}{$#2$}}%
\def\BibTeX{{\rm B\kern-.05em{\sc i\kern-.025em b}\kern-.08em
T\kern-.1667em\lower.7ex\hbox{E}\kern-.125emX}}
\usepackage{balance}
\begin{document}

{\setlength{\fboxsep}{0pt}
\setlength{\fboxrule}{0pt}
}
\title{A Compact Optical Six-Axis Force/Torque Sensor for Legged Robots Using a Polymorphic Calibration Method}

\author{Hyun-Bin Kim$^1$, Keun-Ha Choi$^2$,and Kyung-Soo Kim$^1$,
\emph{Member,~IEEE}\thanks{Manuscript created March, 2023; This work was developed by the MSC (Mechatronics, Systems and Control) lab in the KAIST(Korea Advanced Institute of Science and Technology which is in the Daehak-Ro 291, Daejeon, South Korea(e-mail: youfree22@kaist.ac.kr; choiha99@kaist.ac.kr; 
kyungsookim@kaist.ac.kr).  \emph{(Corresponding
author: Keun-Ha Choi and 
Kyung-Soo Kim)}.}}


\markboth{}
{A Compact Optical Six-Axis Force/Torque Sensor for Legged Robots Using a Polymorphic Calibration Method}

\maketitle

\begin{abstract}
This paper presents a novel design for a compact, lightweight 6-axis force/torque sensor intended for use in legged robots. The design promotes easy manufacturing and cost reduction, while introducing innovative calibration methods that simplify the calibration process and minimize effort. The sensor's advantages are achieved by streamlining the structure for durability, implementing noncontact sensors, and providing a wider sensing range compared to commercial sensors. To maintain a simple structure, the paper proposes a force sensing scheme using photocouplers where the sensing elements are aligned in-plane. This strategy enables all sensing elements to be fabricated on a single printed circuit board, eliminating manual labor tasks such as bonding and coating the sensing elements. The prototype sensor contains only four parts, costs less than \$250, and exhibits high response frequency and performance. Traditional calibration methods present challenges, such as the need for specialized equipment and extensive labor. To facilitate easy calibration without the need for specialized equipment, a new method using optimal control is proposed. To verify the feasibility of these ideas, a prototype six-axis F/T sensor was manufactured. Its performance was evaluated and compared to a reference F/T sensor and previous calibration methods.
\end{abstract}


\section{Introduction}

\IEEEPARstart{R}{ecently,} robots such as Boston Dynamics' Spot, MIT Cheetah3, Mini Cheetah, and ETH's ANYmal have become famous and are being commonly used in a variety of situations. Additionally, manipulator robots that can cooperate with humans are used in factories. Thus, the safety of the robots has become crucial. To ensure safety, robots need force torque (F/T) sensors that can accurately measure force~\cite{ref2,ref3,ref4}.
The requirements for force sensors vary depending on their application. For surgical robots, there is a need for fast response times and high accuracy~\cite{ref5}. The size of the sensor and its load capacity also play significant roles in this context. For quadrupedal robots, sensors should be small in size, exhibit fast response times, and demonstrate impact durability due to dynamic situations~\cite{ref6,ref7}.

In this field, lightweight and impact-resistant force torque sensors are required due to the assumptions of masslessness for the robots' legs and the frequent impacts occurring during walking. Force sensors that employ the strain gauge method are susceptible to impact damage. Impact forces directly transmit to the sensors attached to the spring structure, often causing breakdowns of strain gauge force sensors in such applications. Consequently, commercial walking robots typically utilize one-axis pressure sensors, like Force Sensing Resistors (FSR), rather than multi-axis force/torque sensors in their feet. For instance, bipedal robots in the DARPA Challenge utilized force/torque sensors in their feet, but they were unable to walk for extended periods due to sensor failures.

The miniaturization of F/T sensors is particularly crucial for quadrupedal robots. However, miniaturization often results in a reduction in the sensor's measuring range for the sake of resolution. An F/T sensor should be capable of measuring 3 to 5 times the generalized ground reaction forces of legged robots due to the impact experienced during walking. As such, new methods of F/T sensors utilizing capacity, magnetic fields, and optics have been recently studied. Yet, these methods often demonstrate low performance. Additionally, sensors employing strain gauges experience significant performance degradation in the absence of external devices, such as a 16-bit Data Acquisition system (DAQ). Therefore, there is a pressing need for a novel F/T sensor design.

\subsection{Related Works of Force/Torque Sensor for Legged Robots}
In F/T sensors, strain gauges are usually used because of their accuracy. In commercial applications, ATI-ia's F/T sensors are commonly used. At least 12 strain gauges are used in a 6-axis F/T sensor because each strain gauge can measure only one direction of strain. Much research on strain gauge-type sensors has been performed. This research is mainly focused on fabricating a structure that enables crosstalk-free structures and precise measurements based on structural analysis of springs~\cite{ref9}. Additionally, research using 24 or 36 strain gauges to reduce crosstalk has been performed~\cite{ref10,ref18}.

\begin{figure*}[!t]
\centering
\includegraphics[width=2\columnwidth]{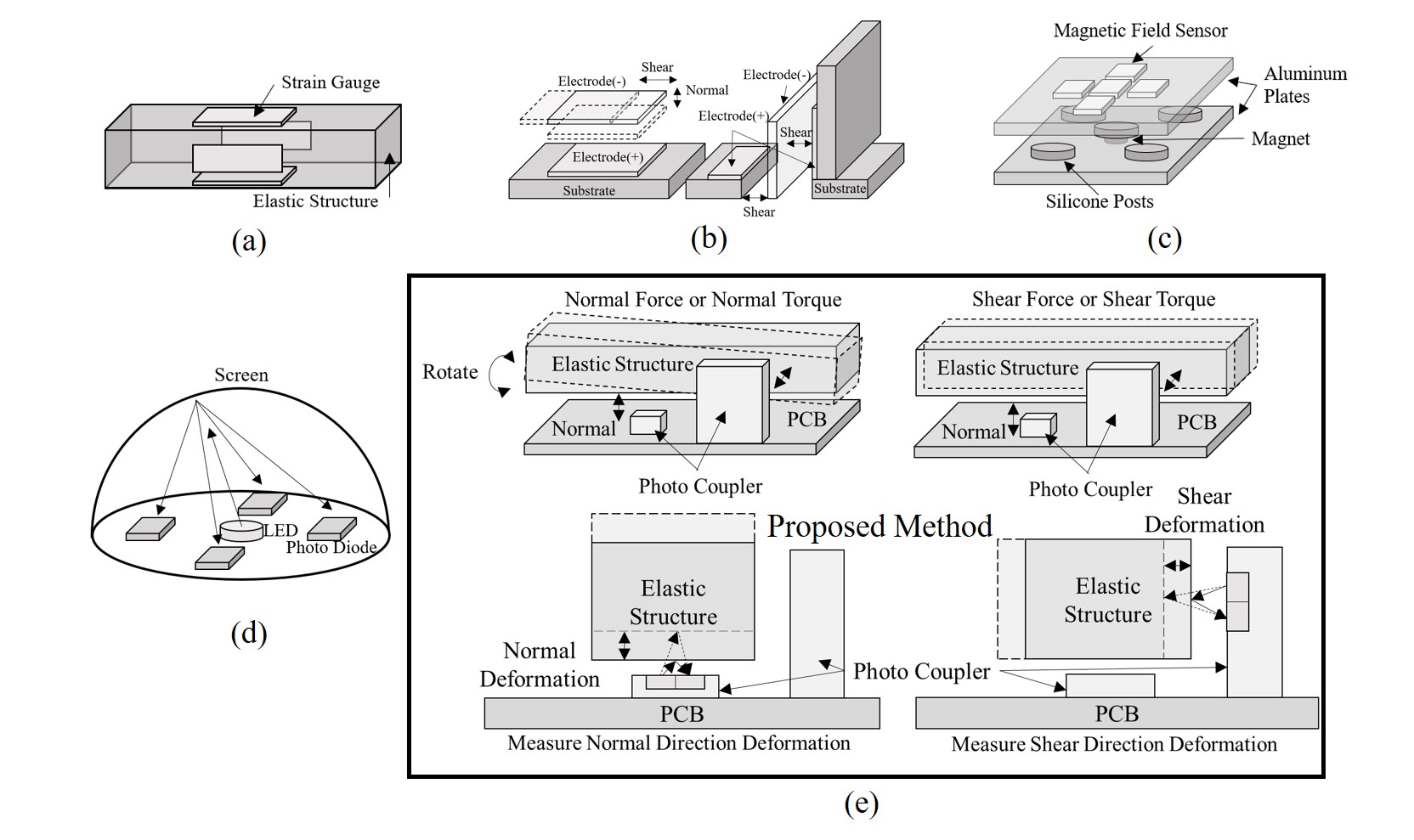}
\caption{Schematics for multiaxis force torque sensors (a)  Strain Gauge Method: This method is employed by the majority of sensors, where strain gauges are affixed to silicon on the surface of an elastic structure. If a force is applied repeatedly, it results in the accumulation of fatigue on the surface where the sensor is attached.
(b) Capacitive Method~\cite{ref13}: This method also involves measuring shear and normal deformation. However, it is susceptible to static electricity if not properly insulated. 
(c) Magnetic field method~\cite{ref14}: This method's resolution is quite low, but it is sensitive in ferromagnetic fields. 
(d) Optic method~\cite{ref16}: This method exhibits robustness upon impact. However, the sensor's nonlinearity is high, necessitating calibration using an Artificial Neural Network (ANN).
(e) Proposed method: In this method, both horizontal and vertical sensors measure normal and shear deformations. The noncontact feature of this method increases the sensor's robustness.}
\label{fig1}
\end{figure*}

In industry, ATI-ia and ETH's Botasys implement strain gauges. Fig.~\ref{fig1} (a) shows the brief operating principle. For example, MINI 85 has 635 g(only transducer) and is divided into a transducer and signal processor. It has a 16-bit resolution system and a wide range of force sensing. ATI force torque sensors are very often used as reference sensors in research because of their accuracy. Botasys's compact sized F/T sensor has a weight of 110 g and a 100 Hz noise-free sample rate. Moreover, new industry and robotics research centers are entering the force sensor market.

For other robot applications, Kim \textit{et al.}~\cite{ref13} made a capacitor-based sensor. Fig.~\ref{fig1} (b) shows the principle of the capacitive method. Based on the characteristics of the capacitor in a small size sensor, it has low sample rates of approximately 200 Hz. This sensor has 10-bit $\sim$ 12-bit resolution. The internal sensor measures noncontact signals and thus has good durability. Additionally, it has low cost~\cite{ref13}.

Additionally, there is a force sensor using magnets made by Ananthanarayanan and Foong~\cite{ref14}, as shown in Fig. \ref{fig1} (c) ~\cite{ref14}. A total of 5 magnet sensors are used to measure the force in the x-axis and z-axis. This also incorporates the principle of measuring force via the displacement that occurs when a force is applied. This sensor has a disadvantage that noise occurs in ferromagnet environments.
Therefore, Epstein~\textit{et al.}~\cite{ref15} made a sensor using piezoresistive elements with a total of 9 sensors. The sensor can measure the force in the x, y and z-axes. Since it is made using flexible silicon and has large nonlinearity, an artificial neural network (ANN) is used for calibration. Although the ANN is used, the maximum error remains at approximately 30\%, which is a large.
The bandwidth of the piezoresistive sensor is up to 1000 Hz. The resolution of the sensor is 12 bits~\cite{ref15}.

For the another robot applications, the On robot (Opto-Force) F/T Sensor uses an optical sensing component. As mentioned in Fig.~\ref{fig1} (d), light emits from one LED in the middle, and the hemispherical spring reflects the light. This reflected light enters the sensor, making it possible for the sensor to measure the 6-axis force and torque through 4 sensors. This sensor has a hemispherical spring that is also a screen, and it shows nonlinear movement because of the hemispherical structure. Since the spring does not contact sensors, it may be more robust. It weighs approximately 350 g, which is heavier than others, and the diameter is 50 mm, which is not small. It has a resolution as low as 9 bits in the case of the z-axis, with the x- and y-axes being 0.5 N and the z-axis is 1 N. The nonlinearity is 2\%, and crosstalk appears within approximately 5\%. It does not outperform the strain gauge method. However, it has the advantage of being strong against impact due to using non-contact method~\cite{ref16,ref28,ref29,ref30,ref31,ref32,ref33,ref34,ref35}.

\subsection{Previous Calibration Methods}

In previous studies, two methods were very often used to calibrate a sensor. One is a method of giving an accurate force for each axis and fitting it using the corresponding sensor value. The second is a calibration method that uses time series data through pseudoinverses (the least squares method). Recently, many calibration methods using ANNs have emerged.
The first method requires linear and rotary stages to give accurate force and motors to operate the stages. It is possible to apply force manually, but it is difficult to exceed 30 N, and it is not possible to apply accurate and continuous force. While this is an easy calibration method if an accurate force can be applied, its disadvantage is that a precise calibration device is needed ~\cite{ref19,ref20}.

The second calibration method is a method of calculating and applying a calibration matrix by the least square method using time series. This method uses the equation
\begin{equation}
    C\ =\ F\cdot\ S^T\ {(SS^T\ )}^{-1}\
\end{equation}
where C is a calibration matrix, S is the sensor value and F is the reference value. Since the use of the Moore-Penrose pseudoinverse (1) is the same as the least square method in a linear system, the Moore-Penrose pseudoinverse is used for calculation~\cite{ref21,ref22}. There is also a supporting method for reducing the nonlinear component by squaring and cubing the value of the sensor using the Taylor series. When calibration is performed using this square of the sensor value, a $6\times12 $ calibration matrix is created, and a more accurate value can be subsequently obtained. However, this method requires many calculations, making calibration difficult, and the number of sensor values to be calculated in the MCU (micro controller unit) increases. Therefore, a $6\times6$ matrix is most commonly used, assuming that it is linear.

Therewithal, there are many cases of calibration using the ANN. Since an F/T sensor itself does not show strong nonlinearity, a shallow layer is usually used. Many methods have been used, such as a fully connected feedforward net and a decision tree~\cite{ref14,ref16,ref15,ref23}. In addition to ANN, calibration research was done using optimization method by Hendrich and Wasserfall~\cite{ref27},
\begin{equation}
\begin{matrix}    
     &\min\limits_{\mathit{C,L,Q}}{\sum\limits_{\mathit{i}\in \mathit{Cal} }{(\mathit{W}_\mathit{ref,i}-\mathit{W}_\mathit{est,i})^2 +\lambda \cdot \left\| \mathit{C,L,Q} \right\|}}
     \\
     \\
     &\mathit{W}=\left(\mathit{F_x,F_y,F_z,M_x,M_y,M_z}\right)
     \\
     & =\mathit{C}+\mathit{LS}+\mathit{S}^\mathit{T}\mathit{Q}\mathit{S}+\ldots 
\end{matrix}
\end{equation}
(2) shows calibration method using cost function as sum of an estimation error and a weighted regularization. 
\subsection{Contributions}
This article has following contributions.
\subsubsection{}
The proposed sensor overcomes the drawbacks of traditional strain gauge-based sensors by utilizing a non-contact sensing approach, resulting in a more robust design against repeated impacts. The sensor employs a spring structure, which has been widely used in the past, for the sake of versatility. This allows for the application of the same principle across various spring structures, providing the advantage of easy implementation and performance consistency. Even when applied to previous structures, the sensor has demonstrated good performance, making it easily adaptable.
\subsubsection{}
Furthermore, the use of the spring itself as the sensing element eliminates the need to attach sensors to the spring, enabling integration on a single printed circuit board (PCB). This reduction in the number of components and labor required for assembly is significant. 
\subsubsection{}
Traditional force torque sensors typically consist of 5 to 7 or more components and employ a larger number of sensors. However, by utilizing only 6 sensors and reducing the number of analog-to-digital converters (ADCs), the proposed sensor reduces the overall component count and allows for the incorporation of additional functionalities. With a total of 4 components, including the PCB, the sensor demonstrates a higher level of robustness compared to designs with a larger number of components. Additionally, it exhibits high sampling rates and accuracy.
In summary, the proposed sensor design addresses the limitations of traditional strain gauge-based sensors by employing a non-contact approach and utilizing a spring structure. The integration of the sensing element and reduced component count results in improved robustness, high sampling rates, and enhanced accuracy.

\subsubsection{}
Also, during calibration, the method of applying force to only one axis requires a device that can apply accurate force separately. Using a pseudo inverse makes it difficult to reflect the null space part and the structural part. The null space can be occurred when the sufficient data is not given. For example, $F_x$ calibration matrix should be [1 0 1 0 0 0], but due to null space, [2 0 -1 0 0 0] or [1 1 1 1 0 -2] can be calculated. To solve this null space problem, $2^{6}$ data (existence and nonexistence of $F_x$, $F_y$, $F_z$, $M_x$, $M_y$, $M_z$) are needed for calibration. 
By utilizing the structural constraints between the spring and the sensors and employing quadratic programming to transform it into an optimal problem, the influence of the null space was minimized, resulting in more accurate and efficient calibration.
\begin{figure}[!t]
\centering
\includegraphics[width=0.9\columnwidth]{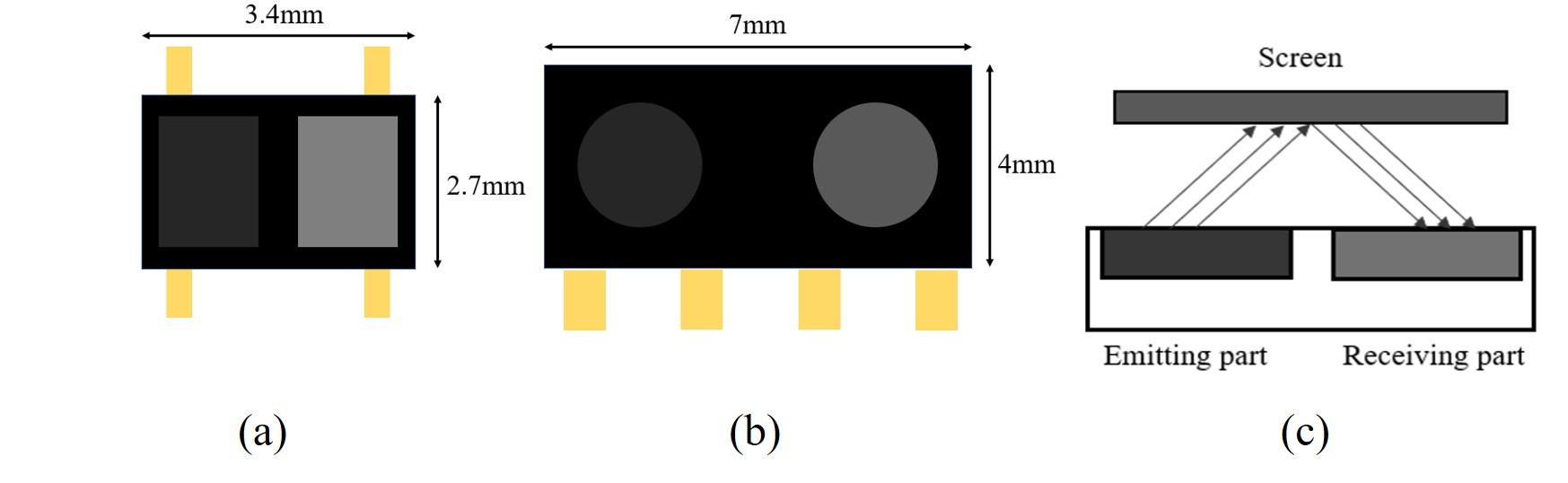}
\caption{Photocouplers Utilized: The compact size of the photocoupler offers a significant advantage when it comes to miniaturizing the sensor. (a) Onsemi-QRE1113GR (b) Vishay-TCRT1000 (c) Principle of the photocoupler.
}
\label{fig2}
\end{figure}
\begin{figure}[!t]
\centering
\includegraphics[width=0.9\columnwidth]{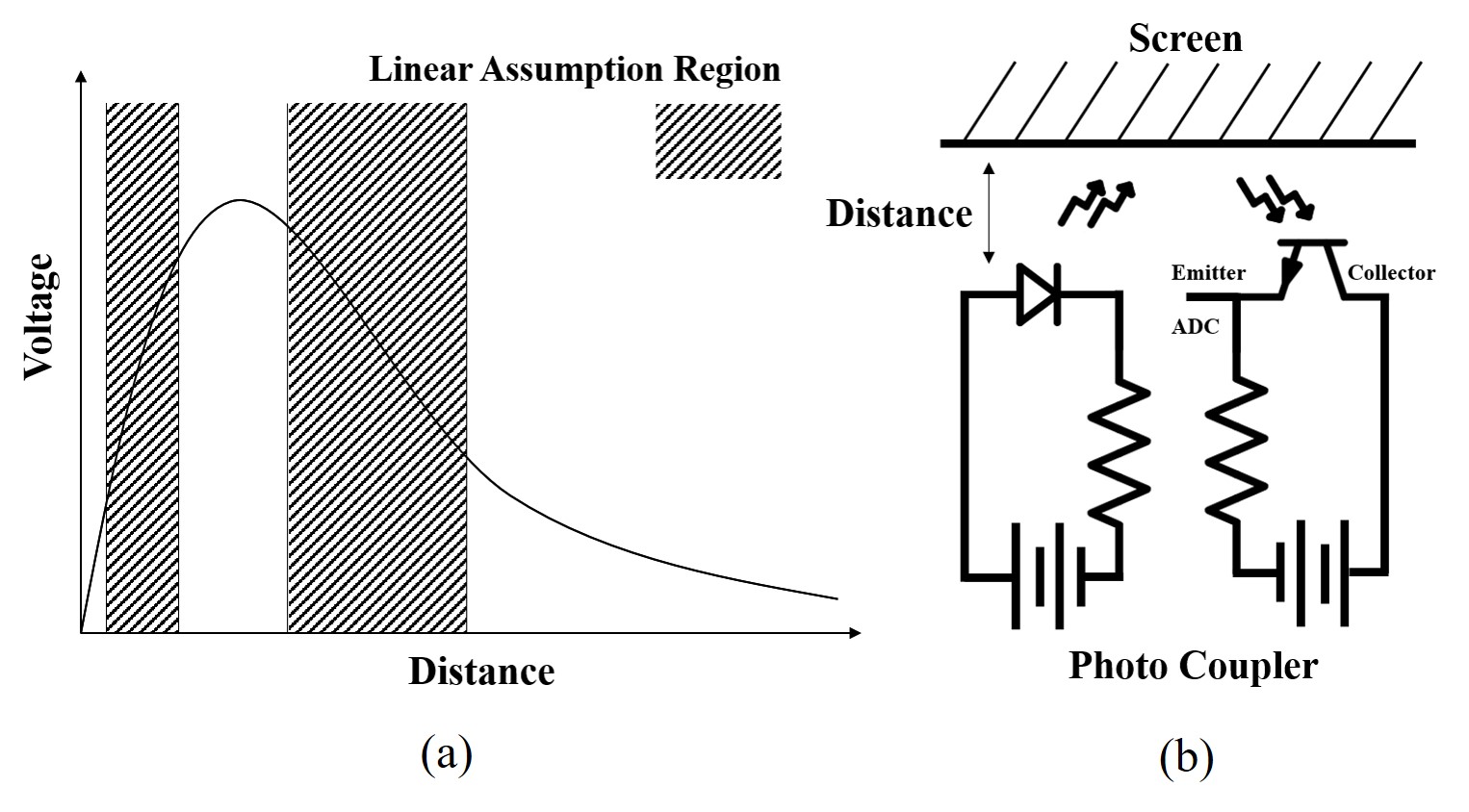}
\caption{(a) Graph Between Voltage and Distance: Upon overall observation, the relationship can be considered nonlinear, but it approximates a linear relationship within certain specific ranges. This is due to the fact that the measured distance range is, at most, approximately 0.1mm.
(b) Photocoupler Circuit: The photocoupler circuit can operate using only two resistors, which makes it simple and cost-effective. To minimize the effect of variations in junction temperature, we employed temperature-insensitive resistors that are commonly used in automotive applications.
}
\label{fig3}
\end{figure}
\section{Sensor Design and Fabrication}
\begin{figure*}[!t]
\centering
\includegraphics[width=2\columnwidth]{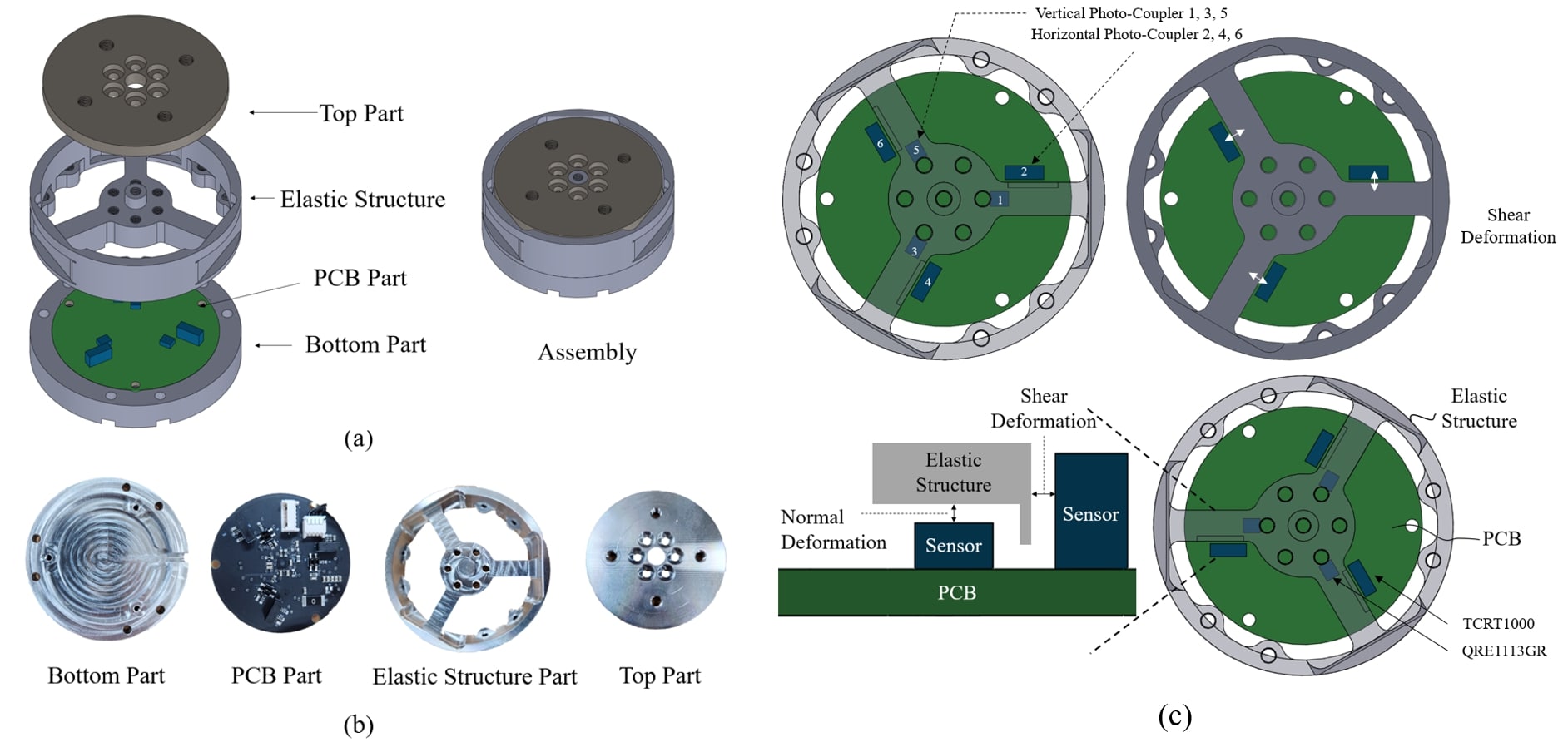}
\caption{(a) 3D Modeling of the Proposed Sensor
(b) Actual Appearance of the Proposed Sensor: The design includes only four parts. The bottom and top parts function collaboratively, serving as the mounting components responsible for applying force.
(c) Sensor Locations and Measurement of Normal and Shear Deformation: In the diagram, the numbers on the sensors correspond to the sensor numbers in TABLE~\ref{tab1}. This diagram illustrates the decomposition of 6-axis forces into horizontal and vertical forces for decoupling purposes. It shows how sensors measure deformations in the horizontal and vertical directions independently.
}
\label{fig4}
\end{figure*}

Photocouplers are commonly used in circuits such as switches in machines and can be composed of small and simple circuits such as end switches. They are widely used in many electronic devices. Because it uses light, it has a fast response, so it goes into high-speed communication isolation circuits, and it is often used in circuits that implement different voltages. Because of its simple structure, it is inexpensive and is most often used for switches. 

However, the photocoupler can not only be used as a digital device but also as an analog device. A photointerrupter or photoreflector has a linear response between screen displacement and voltage, and a sensor can therefore be made with these components.
In this paper, a photoreflector, which is a type of photocoupler types, is used to reduce sensitivity to mechanical tolerances and propose a new type of 6-axis force torque sensor.

The QRE1113GR is small: 3.6 mm and 2.9 mm in size, as shown in Fig.~\ref{fig2}. Due to the small size of sensors, a small device package can be manufactured. Additionally, the phototransistor is inside the device, so no other devices, such as Wheatstone bridges or amplifiers, are needed. The photoreflector usually has two linearly assumable states for measuring distance~\cite{ref33}.

Fig.~\ref{fig3} (a) shows the nonlinearity of graphs between displacement and current. In the case of QRE1113GR, the 0 mm $\sim$ 0.4 mm section and 0.6 mm $\sim$ 1.8 mm section have a linear function between distance and voltage. Since a small displacement of 0.4 mm or 1.2mm can be measured in a full range, it can be applied to a force torque sensor taking advantage of these properties. Fig. ~\ref{fig3} (b) explains the photoreflectors circuit and the application circuit. It has a very simple circuit that can perform sensing and reduce cost. The photocoupler has a fast response time of more than 5 kHz and uses a noncontact sensor measurement method. By way of the noncontact measurement, the durability of the sensor increases to beyond that of the strain gauge method.

\begin{figure}[!t]
\centering
\includegraphics[width=0.7\columnwidth]{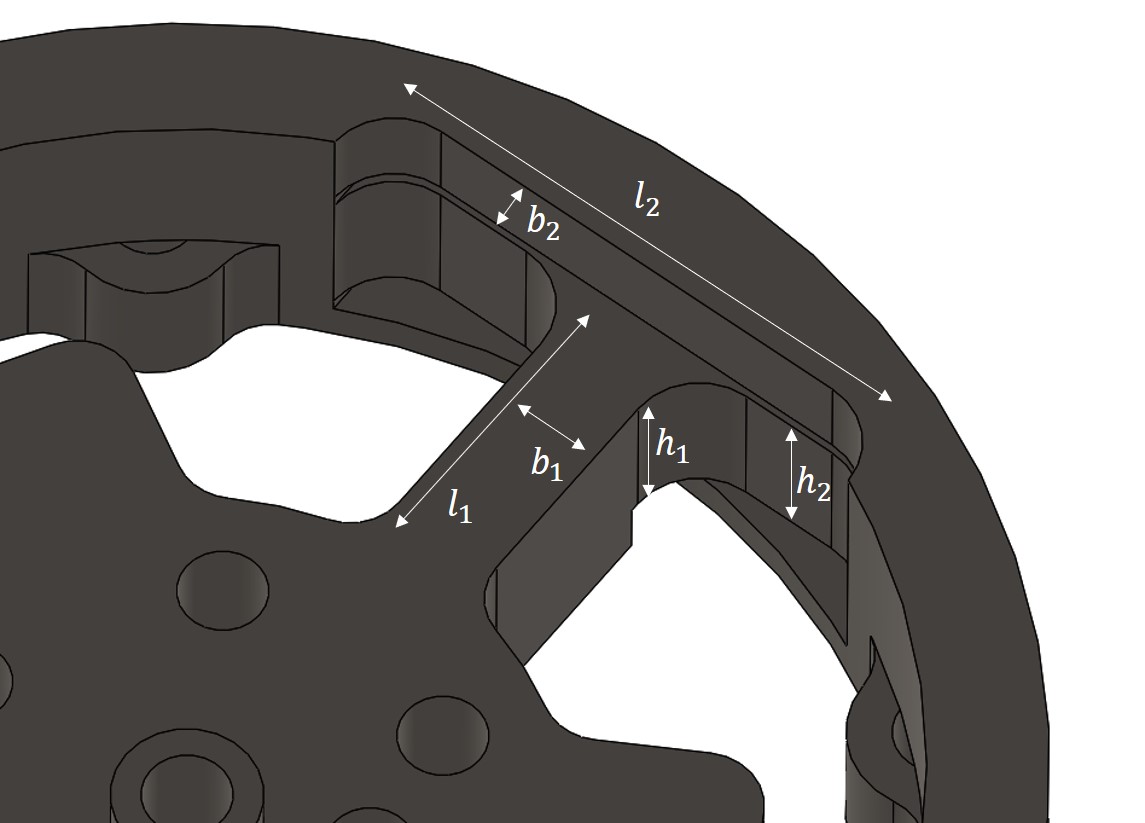}
\caption{Cross Elastic Beam Structure: The values are listed in TABLE~\ref{tab2}. They were determined by measuring the range for further research application, with the help of SolidWorks FEM, as described in the Appendix.
}
\label{fig5}
\end{figure}
\begin{table}[!t]
\begin{center}
\caption{Sensor Values Applying Each-Axis Force and Torque}
\label{tab1}
\begin{tabular}{c c c c c c c}
\hline\hline
     & Sensor1     & Sensor2     & Sensor3     & Sensor4     & Sensor5     & Sensor6     \\ \hline
$F_x$ & $\sim$ & $\sim$ & $\sim$ & $++$ & $\sim$ & $--$ \\
$F_y$   &     $\sim$          &    $++$          &      $\sim$         &     $-$         &          $\sim$     &    $+$          \\
$F_z$    &         $-$     &      $\sim$         &       $-$       &         $\sim$      &        $-$      &    $\sim$           \\
$M_x$    &         $\sim$      &       $\sim$        &        $++$      &       $\sim$        &      $--$        &         $\sim$      \\
$M_y$    &        $++$      &        $\sim$       &     $-$         &      $-$        &        $\sim$       &         $\sim$      \\
$M_z$    &           $\sim$    &           $+$   &           $\sim$    &   $+$           &              $\sim$ &       $+$       \\ \hline
\end{tabular}
\end{center}
\end{table}

Figs.~\ref{fig4} (a, b) shows that the prototype sensor consists of 4 parts: a bottom part, an elastic structure part, a top part, and a printed circuit board.
To fabricate the sensor without labor using only one printed circuit board, sensors must be placed on a plate. The following theory was used to measure the 6-axis forces.

Fig.~\ref{fig4} (c) shows the screen position and shape.
There are six screens. The three screens are for normal-axis forces ($F_z$, $M_x$, $M_y$), and others are for shear-axis forces ($F_x$, $F_y$, $M_z$).

The transfer function can be expressed as (2)
\begin{equation}
    X=AF
\end{equation}
where $X\in \mathbb{R}^6$ is a displacement due to force, $F\in \mathbb{R}^6$ is the measured force and $A\in \mathbb{R}^{6\times6}$ is the matrix between the displacement and the force.
In TABLE~\ref{tab1}, one plus sign($+$) indicates that the voltage variation is slightly high, and double plus sign($++$) indicates the voltage variation is higher than one plus sign($+$). Additionally, the minus sign ($-$) is the opposite of the plus sign ($+$). A tilde ($\sim$) indicates that there is very little variation or no variation. There are 6-axis forces, $F_x$, $F_y$, $F_z$, $M_x$, $M_y$, $M_z$ and 6 screens' results in this table. In $F_n$, $F_n$ is the force that is the $n$ axial force. In $M_n$, $M_n$ indicates the moment that is the $n$ axial moment.
From the rank of this matrix, if this matrix is full rank, 6 forces can be measured by 6 sensors by using linear algebra theory.
Experiments confirmed this design from the SolidWorks FEM based on force and displacement results. The result of the matrix shows that it has full rank. This signifies that 6 sensors can measure the 6-axis forces~\cite{ref26}.

The elastic structure in Fig.~\ref{fig5} can be calculated by Timoshenko beam theory. The parameters of the cross elastic beam are shown in TABLE~\ref{tab2}.
It has 6 parameters, which were decided by the target F/T measuring range that can be used in the feet of my quadruped robot. Parameters could be calculated with SolidWorks FEM.

\begin{figure*}[!t]
\centering
\includegraphics[width=1.8\columnwidth]{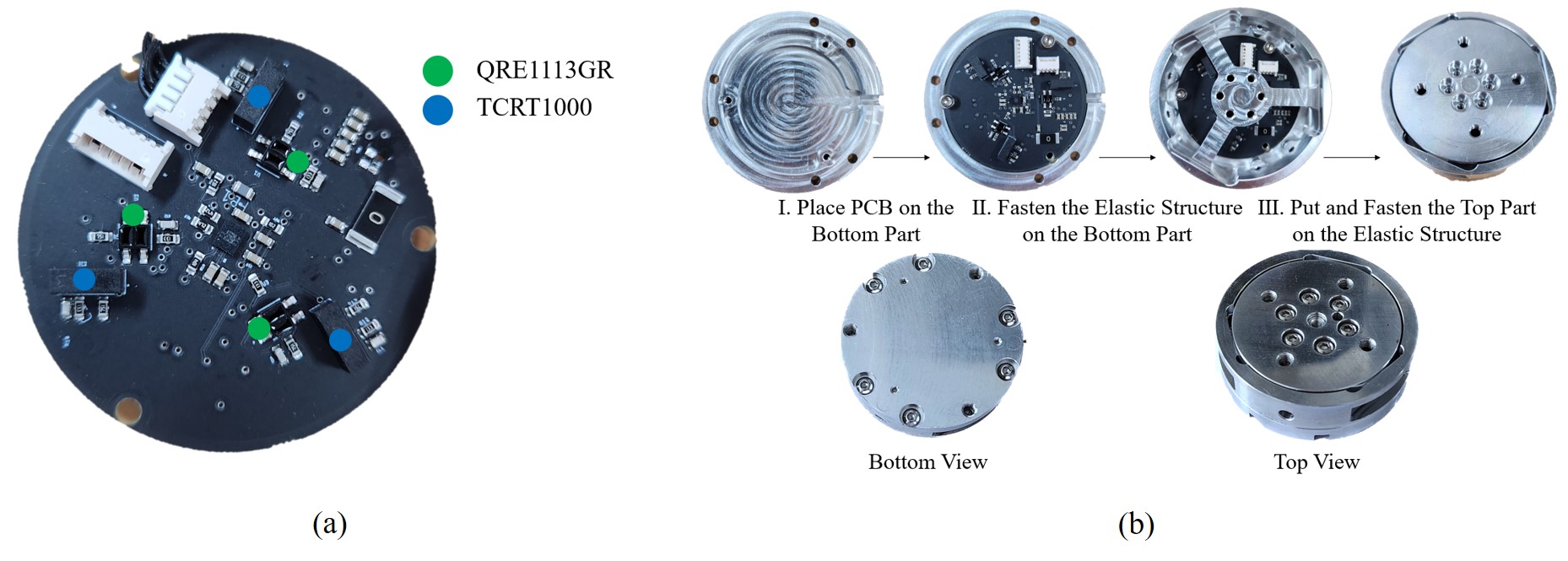}
\caption{(a) Printed Circuit Board and Electronic Circuit of the Proposed Sensor: The QRE1113GR (marked by a green circle) is oriented upwards, which allows it to measure vertical forces. The TCRT1000 (marked by a blue circle) is oriented sideways, enabling it to measure horizontal forces. In the center of the PCB, there is an IMU. On the opposite side of the PCB, there is an MCU which converts the sensor values into forces.
(b) Simple Manufacturing Process of the Proposed Sensor: The manufacturing process of the proposed sensor is straightforward and consists of only three steps - placing the PCB and spring, and then fastening them with bolts.
}
\label{fig6}
\end{figure*}
\begin{table}
\begin{center}
\caption{Mechanical Specification of Prototype Sensor}
\label{tab3}
\begin{tabular}{c c c c c c c}
\hline\hline
  &Value    &Unit       \\ \hline
  Diameter&56&mm\\
  Height&       18&     mm\\
Material&       AL7075-T6  & \\
Input Force Range X     &$\pm$1,050&    N \\
Input Force Range Y     &$\pm$1,200&    N \\
Input Force Range Z     &$\pm$1,850&    N \\
Input Moment Range X, Y &$\pm$25        &N$\cdot$m \\
Input Moment Range Z    &$\pm$36&       N$\cdot$m \\
Sampling Frequency      &5&     kHz \\
Weight  &79&    g \\
Resonance Frequency X &12006&Hz\\
Resonance Frequency Y &12039&Hz\\
Resonance Frequency Z &14154&Hz\\
 \hline
\end{tabular}
\end{center}
\end{table}
The parameters of the elastic structure can be easily customized, allowing for simple adjustments to the sensor's F/T measuring range.
According to our previous experiments, the sensor design operates effectively in a plane. The printed circuit board (PCB) consists of three main components. First, 6 photocouplers were positioned similarly to the arrangement previously described. Second, two types of heat-insensitive resistors were included for LEDs and transistors. Third, an Analog-to-Digital Converter (ADC) was connected. These three components are all that are required to construct the sensor, contributing to its ease of fabrication and low cost. Fig.~\ref{fig6} (a) presents the PCB design, which is straightforward overall. A STM32H7 Micro Controller Unit (MCU), Controller Area Network (CAN) transceiver, and Inertial Measurement Unit (IMU) are attached to the PCB, allowing force calculations to be performed by the MCU and communication with a PC or other devices via CAN. Consequently, the proposed sensor only requires four connections, including power and CAN (H/L), eliminating the need for a separate DAQ and facilitating a compact design.

The mechanical components of the F/T sensor consist of only 3 parts. The top part is designed to attach the actuator, which applies force. The elastic structure is a frame with a screen for displacement due to force, similar to a spring. The bottom part is meant to attach to the force-receiving component. Fig.~\ref{fig6} (b) illustrates the simplicity of its manufacturing process. Assembling the entire sensor requires just three steps and only involves bolting.
The sensor is compact in both diameter and height, and it is constructed from AL7075-T6 due to the material's low silicon content and minimal hysteresis. The sensor can handle a maximum input force of 1850 N for the z-axis. A key advantage of the photocoupler is its versatility - the sensor can be used across all sections as broadly as the mechanical properties allow.
TABLE~\ref{tab3} lists the mechanical specifications of the prototype sensor. It boasts an extensive range for measuring forces and moments, and it is remarkably lightweight at just 79 g, including the weight of the mounting plate and cable. The resonance frequency was calculated as 12000Hz, making it sufficient for measuring high-frequency forces.

\section{Experiment Setup}
\subsection{The Basic Sensor Property and Bandwidth Test}
\begin{figure}[!t]
\centering
\includegraphics[width=0.9\columnwidth]{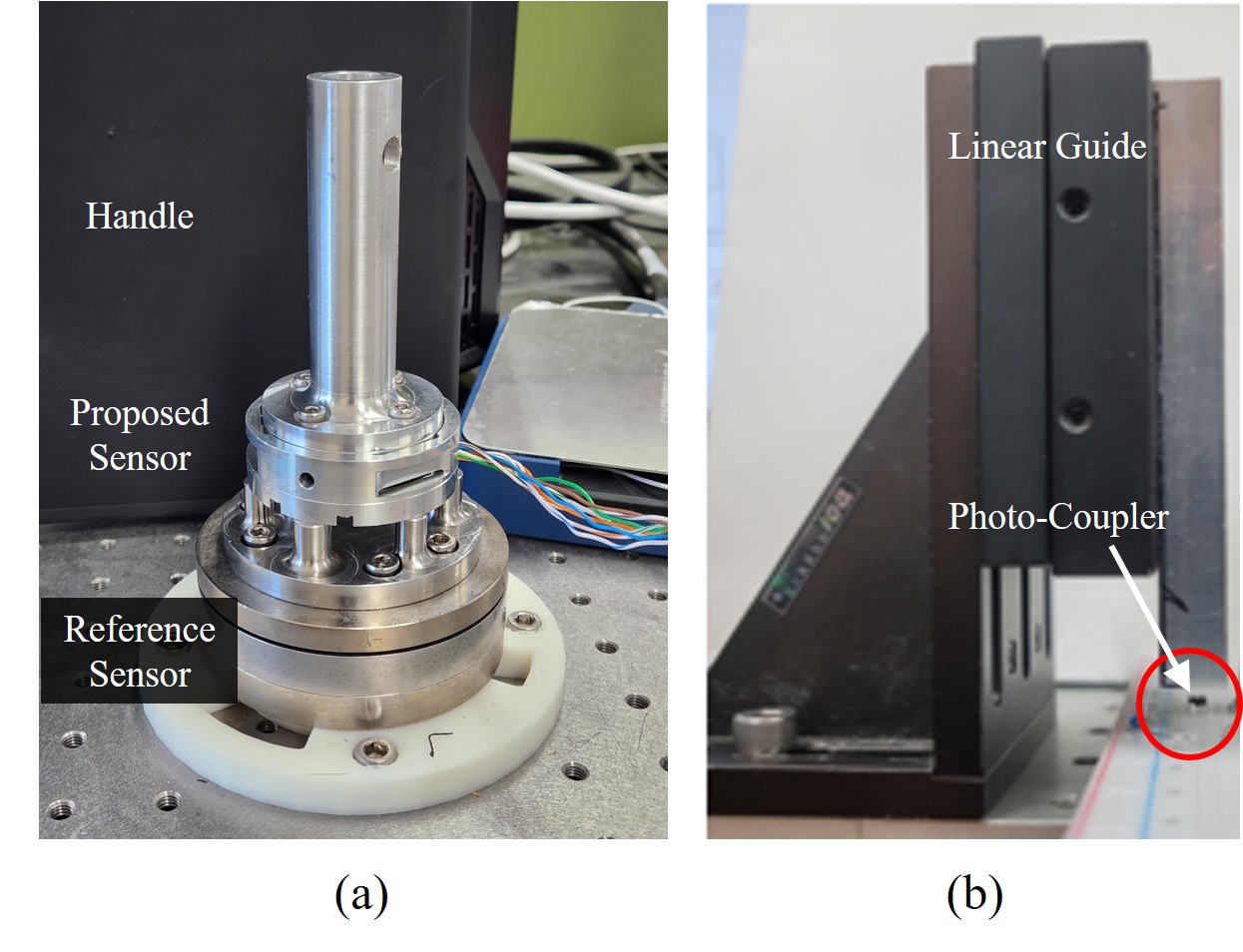}
\caption{Experimental setup used to compare the six-axis F/T measurements taken by the proposed sensor with those from the reference sensor.
(a) Calibration Experiment Environment: The experimental equipment comprised a handle, the proposed sensor, a mount between the proposed sensor and reference sensor, and the reference sensor itself.
(b) Single Photocoupler Experiment Environment: This experiment used a linear guide to adjust the distance between the screen and the photocoupler sensor.
}
\label{fig8}
\end{figure}

For the prototype sensor's property, the property of each photocoupler has to be tested, when photocoupler's input voltage is 5V. This test was performed using one photocoupler to record the standard deviation and covariance of white noise and frequency response. Kalman filter's covariance value was decided by these values. The sensor's response can be assumed to be linear across 0.0 mm-0.4 mm. Each of the sensor's standard deviation and covariance when the sensor is in the steady state is 3.92$e^{-2}$V and 1.54$e^{-5}$V.
The test for the sensor bandwidth experiment using step input was performed. This experiment was done as giving step input to the LED, and measuring the transistor value at that time(sampling rate : 10kHz, 20kHz). Following Figs. \ref{fig7} (a, b) show the results of the test. Without the Kalman filter, QRE1113 shows 5 kHz bandwidth performance, and TCRT1000 shows 3 kHz bandwidth performance.

\subsection{Calibration Setup}

As a calibration experiment, MINI-85 (ATI Ind, Autom.) was used as the reference F/T sensor that has a 2 percentage full scale error which is quite low error. Additionally, NI-DAQ, PeakCAN USB and MATLAB Simulink were used to perform this experiment. The experiment was designed to compare the response of the prototype photointerrupter sensor with a commercial reference 6-F/T sensor (Mini-85, ATI Ind. Autom.). Fig.~\ref{fig8} shows the experimental setup. The commercial reference sensor and the prototype photocoupler F/T sensor are attached serially. The experiment was performed by applying force and torque at the over the top part of the prototype sensor. After applying force and torque at the sensors, a matrix that can calibrate between the measured value from the prototype sensor and the real force value can be calculated. Three calibration methods were used: the proposed method, neural network and least square method.

\subsection{Calibration Method}
\begin{table}
\begin{center}
\caption{Ratio of Distance Changes by Each-Axis Force and Torque(FEM)}
\label{tab5}
\begin{tabular}{c c c c c c c}
\hline\hline
     & Sensor1     & Sensor2     & Sensor3     & Sensor4     & Sensor5     & Sensor6     \\ \hline
$F_x$ &0.03&-0.09&-4.29&119.30&4.32&-116.79 \\
$F_y$ &5.01&-117.80&-2.41&59.10&-2.44&82.62  \\
$F_z$ &124.60&-0.79&124.60&-0.34&124.60&-4.92   \\
$M_x$ &15.76&-41.04&-81.90&18.96&67.66&17.37   \\
$M_y$ &88.10&-0.70&-27.98&-32.13&-54.08&33.15    \\
$M_z$  &-4.23&70.82&-4.23&70.83&-4.26&71.00    \\ \hline
\end{tabular}
\end{center}
\end{table}
Quadratic programming is a process that can solve mathematical optimization problems by using quadratic functions. Quadratic programming is used to find an n-dimensional vector x that satisfies the following Equation (4). \begin{equation}
    \begin{matrix}\mathit{minimize}\ \frac{\mathbf{1}}{\mathbf{2}}
    \mathit{x}^\mathit{T}\mathit{Qx}+\mathit{c}^\mathit{T}\mathit{x}\\
    \begin{matrix}
    \mathit{subject}\ \mathit{to}\ \mathit{Ax}\le\mathit{b},\\
    \mathit{Ex}=\mathit{d}\\
    \end{matrix}\\
    \end{matrix}
\end{equation}
,where $x$ represents the state of the system, while $Q$ and $c$ are related to the cost of the optimization problem. $A$ denotes the matrix for inequality constraints, and $b$ is the constant for these constraints. Similarly, $E$ is the equality constraint matrix, and $d$ is the constant for these equality constraints.
Quadratic programming incorporates both equality and inequality constraints. However, only inequality constraints were utilized in this study.
In recent times, this method has been widely used due to advancements in computational capabilities. This paper introduces a new calibration method for a 6-axis F/T sensor utilizing quadratic programming. Prior research typically used least squares methods for sensor calibration.
Nonetheless, the least squares method poses a significant challenge, as it may not apply appropriately to the structural relationship of the sensor. Consequently, calibration issues may arise when employing the least squares method.
The 6-axis F/T sensor has a calibration matrix $\mathbb{R}^{6\times6}$, which should have a rank of 6. However, each axis has 5 null spaces which lead to an infinite number of solutions in calibration. To address this issue within the framework of quadratic programming, this study employs an inequality that is capable of encapsulating the structural mathematical relationship.

TABLE~\ref{tab5} presents the variation in sensor values corresponding to the forces and torques on each axis. These values are derived using SolidWorks FEM (Finite Element Method). They illustrate the relationship between the screen displacement and the applied force for each axis. In particular, every sensor exhibits its own unique relationship for each axis, as detailed in TABLE~\ref{tab1}.

\

\begin{equation}
 \begin{matrix}   F_x:&\ S_4+S_6=0\\
F_y:&\ S_2+S_4+S_6=0,\ S_4=S_2 \cdot a_1,\\&\ S_6=S_2 \\
F_z:&\ S_1=S_3=S_5\\
M_x:&\ S_3+S_5=0\\
M_y:&\ S_1+S_3+S_5=0,\ S_1=S_5=-a_2\cdot S_3\\
M_z:&\ S_2=S_4=S_6\\
\end{matrix}
\end{equation}.
In the provided formulas, $F_n$ represents the force on the nth axis, $M_n$ stands for the moment on the nth axis, and $S_n$ denotes the value of the nth sensor. $a_1$ and $a_2$ are constants can be calculated by geometric structure such as $cos(\pi/3)$. The sensors $S_1$, $S_3$, and $S_5$ are oriented vertically, while sensors $S_2$, $S_4$, and $S_6$ are positioned horizontally. Equation (5) can be derived based on the geometrical structure illustrated in Fig.~\ref{fig4}(c), which also serves as the constraint for the sensor calibration matrix.

Constraint (5) can thus be applied for calibration purposes. In the context of the least squares method, the calibration matrix C can be determined using the Equations (6).


\begin{equation} \begin{matrix}
F&=&\ C \cdot \ s\\
s^+&=&{(s^Ts)}^{-1}s^T  \\
C&=&F{(s^Ts)}^{-1}s^T\\\end{matrix}
\end{equation}
In these equations, $s$ denotes the time series data of sensor ADC values, $C$ represents the calibration matrix, and $F$ signifies the time series data of the reference force and torque. When $s$ and $F$ are known parameters and $C$ is the unknown parameter, many solutions for the calibration matrix C can arise in the absence of specialized equipment to precisely apply force on each axis. This is due to the equation lacking structural constraints.
Calibration can be viewed as a form of system identification, and quadratic programming can serve as an optimal problem-solving approach. By employing the transpose of the time sequential matrix, we can formulate the system identification issue as an optimization problem.
To construct the quadratic programming form, the following Equations (7) are required.



\begin{equation}
    \begin{matrix}
\ h &=&s\\
\ d &=&F_x     \\
\ H &=&2h^Th\\
\ c &=&-2h^Td
    \end{matrix}
\end{equation}.

In these equations, $s$ stands for the time series data of the sensor values, while $F_x$ represents the time series data of the x-axis reference force values. These variables can be used to calculate H and c. This equation effectively transforms the time series data into the costs of the quadratic programming. Subsequently, these can be incorporated into the quadratic programming as shown in the following example. The structural constraints matrix A can be determined by referring to TABLE~\ref{tab5}.


\begin{equation}
\begin{matrix}
    A=\ \left[ \begin{matrix} 231&0&232&0&-28851&0\\
0&8.56&0&8.38&0&0\\
0&-8.56&0&-8.38&0&0\\
0&0&0&-8.38&0&-10689\\
0&0&0&8.38&0&10689
    \end{matrix} \right] \\
\\

    b=\left[ \begin{matrix}100&100&100&100&100 \end{matrix} \right]^T \\
\\

    \min\limits_{x} {\frac{1}{2}}x^THx+c^Tx\ \ such\ \ that\ Ax\le b \\

\end{matrix}
\end{equation}.
In these equations, $x$ represents the calibration matrix for the x-axis force, while $A$ denotes the geometric structure constraint matrix for the x-axis force, and $b$ is a constant for the inequality constraint.

The $A$ matrix is derived from the structural constraints as determined by SolidWorks FEM. The first row of matrix $A$ mirrors the equation $S_4 = S_6$, and the fourth row mirrors the equation $|S_4|>>|S_2|$. The second row reflects the equation $S_3\sim S_5,|S_3|>>|S_1|$. The values for these equations can be calculated using TABLE~\ref{tab5}. In equation (8) above, the first column of matrix $A$ corresponds to Sensor4, while the second to sixth columns correspond to Sensor3, Sensor6, Sensor5, Sensor2 and Sensor1 in TABLE~\ref{tab5}. The second to fifth rows of matrix $A$ represent the relationships between normal force sensors when moments occur due to forces.
$b$ accounts for the matrix error caused by sensor gain and mechanical tolerance. By incorporating these values into the quadratic programming Equations (8), the calibration matrix $x$ can be computed. Given the errors associated with sensor gain and mechanical tolerance, inequality constraints were used instead of equality constraints. The value of $b$ is determined by the gain of the calibration matrix and can be customized and adjusted to enhance the fit. This indicates that manual parameter adjustment can be carried out during calibration, which is not feasible with the pseudo inverse method. The more adjustable factors, the higher the sensor's accuracy.


\begin{figure*}[!t]
\centering
\includegraphics[width=2\columnwidth]{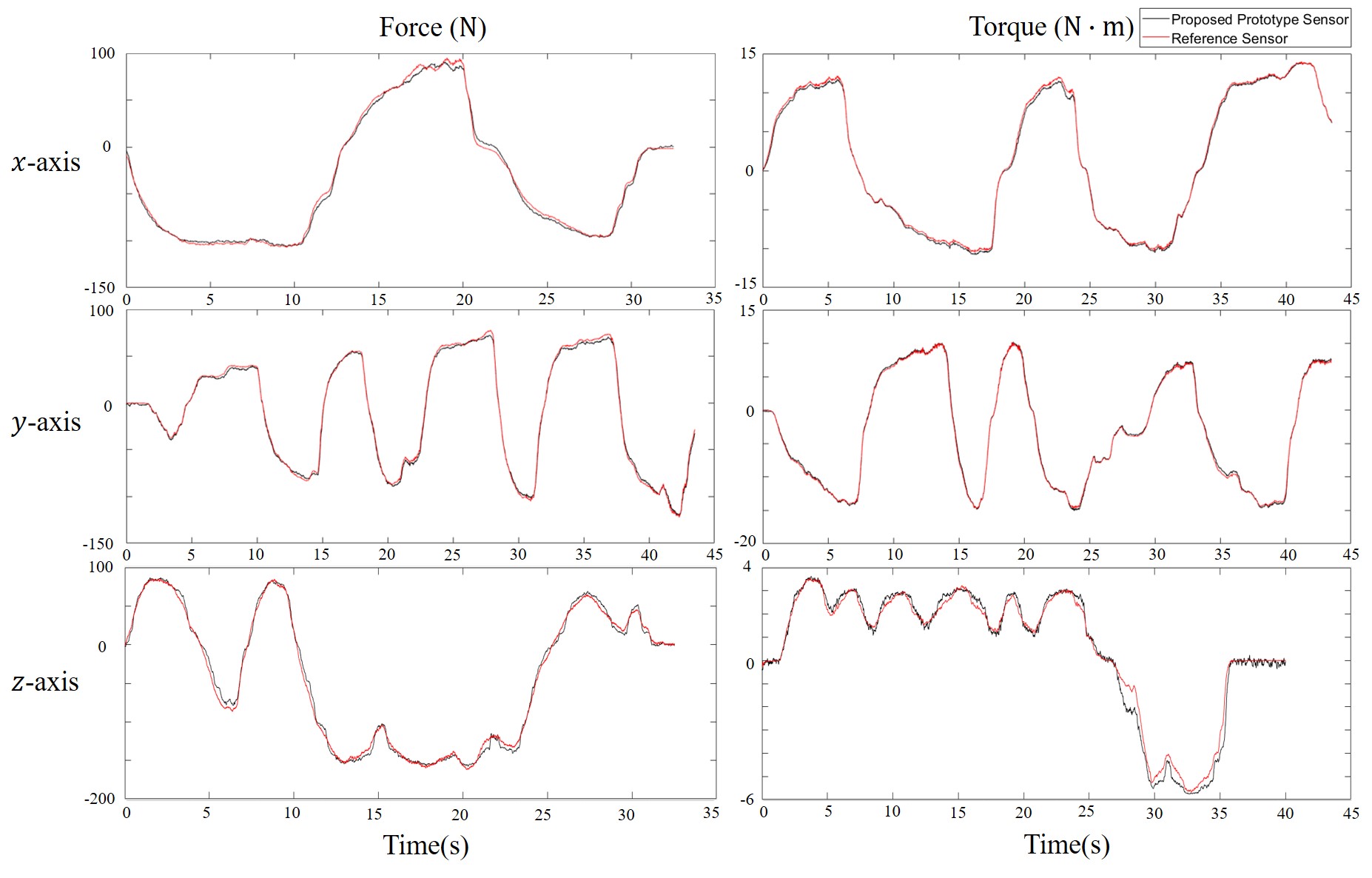}
\caption{Time-Domain Result of the Comparison between the Calibrated Proposed Sensor and the Reference Sensor: The black lines represent the results from the prototype sensor, while the red lines represent the results from the reference sensor (MINI-85).
}
\label{fig9}
\end{figure*}

\section{Experimental Evaluations}
\subsection{Proposed Sensor Calibration Results}
\begin{table}[!t]
\begin{center}
\caption{Full-Scale Percentage Error And Root-Mean-Square Error}
\label{tab6}
\begin{tabular}{c c c c c}
\hline\hline
  & & Percentage Error(\%) &   &\multirow{2}{*}{RMS Error(N, N$\cdot$m)}       \\
  &Mean &Std &Max&  \\ \hline
  \ $ F_x $ &0.13&0.092&0.42&2.72   \\
  \ $ F_y $ &0.085&0.053&0.31&2.25   \\
  \ $ F_z $ &0.096&0.075&0.60&4.52   \\
  \ $ M_x $ &0.51&0.35&2.06&0.31   \\
  \ $ M_y $ &0.40&0.27&1.54&0.24   \\
  \ $ M_z $ &0.38&0.30&1.94&0.35   \\
  \hline
\end{tabular}
\end{center}
\end{table}

\begin{table}[!t]
\begin{center}
\caption{Resolution and Sensitivity}
\label{tab7}
\begin{tabular}{c c c }
\hline\hline
\multirow{4}{*}{Resolution} & Force(X,Y axis)       & 0.05N       \\
&Force(Z axis) & 0.1N\\
                            & Torque(X,Y axis)      & 0.02N$\cdot$m      \\ 
                             & Torque(Z axis)      & 0.05N$\cdot$m      \\ \hline
Sensitivity                 & \multicolumn{2}{c}{0.38mV/N}
 \\ \hline
\end{tabular}
\end{center}
\end{table}
The proposed sensor is evaluated by its time response, full scale error, root-mean-square error, resolution and sensitivity. Fig.~\ref{fig9} shows the time response when comparing force and torque between the reference sensor and the proposed sensor. The result of the proposed sensor response is quite accurate as shown as TABLE~\ref{tab6}. Additionally, this result is calculated in terms of the full-scale percentage error and RMSE (root-mean-square error).
The results show that the z-axis force and torque errors are larger than those of the other axes. The spring structure's z-axis measuring range is larger than that of the other axes, resulting in the z-axis error being larger.
\begin{figure}[!t]
\centering
\includegraphics[width=\columnwidth]{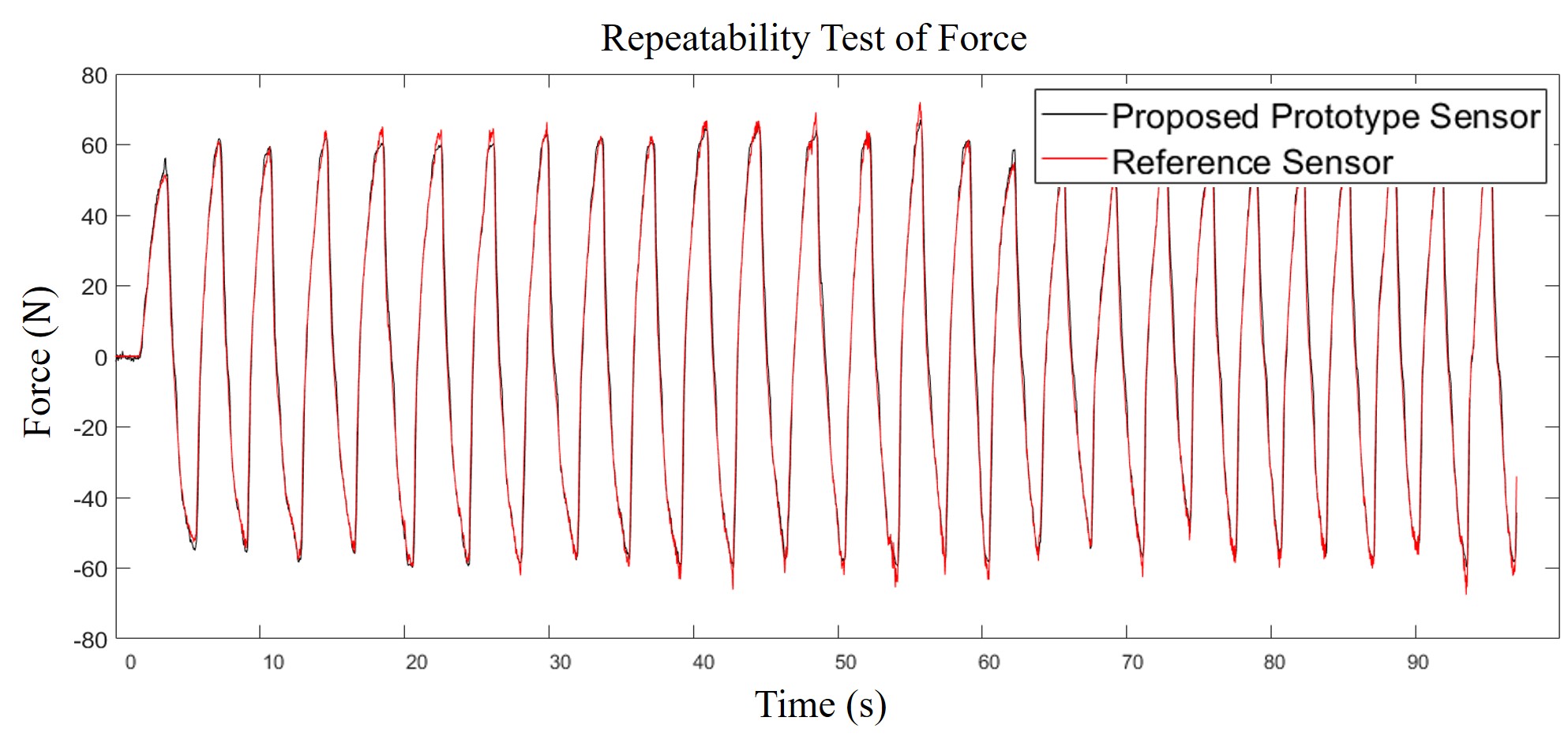}
\caption{Repeatability Test of Force Sensor (x-axis): This test demonstrates the robustness of the proposed sensor. It maintains consistent sensor values even when subjected to repeated application of force, as indicated in the repeatability experiment.
}
\label{fig10}
\end{figure}


The resolution and sensitivity of the proposed sensor are computed as shown in TABLE~\ref{tab7}. The MCU is equipped with 16-bit resolution analog to digital converters that operate within the range of $0$ V to 3.3 V. As displayed in TABLE~\ref{tab7}, the force resolution is roughly 0.1 N, and the torque resolution is approximately 0.05 N$\cdot$m. In digital bits, this resolution can be computed to a maximum of 14$\sim$15 bits. These results demonstrate quite high resolution for F/T sensors that do not have external DAQ systems.

\begin{table}[!t]
\begin{center}
\caption{Cross-talk Error(Full-Scale Error)}
\label{tab8}
\begin{tabular}{c c c c c c c}
\hline\hline
Axis & $F_x$      & $F_y$       & $F_z$       & $M_x$      & $M_y$     & $M_z$       \\ \hline
$F_x$   & N/A     & 2.5\%    & 1.6\% & 1.3\%  & 1.3\% & 0.1\% \\
$F_y$   & 2.5\%   & N/A      & 2.5\%    & 1.0\%     & 1.6\%  & 0.4\% \\
$F_z$   & 1.9\% & 2.5\%    & N/A      & 3.1\% & 0.2\%  & 0.4\% \\
$M_x$   & 1.3\%  & 1.0\%      & 3.1\%  & N/A     & 0.6\%  & 0.3\% \\
$M_y$   & 1.3\%  & 1.6\% & 3.1\%  & 0.8\%   & N/A    & 0.3\% \\
$M_z$   & 3.1\% & 2.5\%    & 1.6\% & 2.0\%     & 2.0\%    & N/A      \\ \hline
\end{tabular}
\end{center}
\end{table}
The cross-talk of the proposed sensor is presented in TABLE~\ref{tab8}. The maximum cross-talk was 3.125\% in terms of full-scale error. The time domain responses of force and moment are displayed in Fig.~\ref{fig20}. These results demonstrate that the performance of the proposed sensor is comparable to that of the reference sensor.

\subsection{Calibration Results}
\begin{equation}
    \begin{matrix}
       \ A_{pse} = \Scale[0.8]{\left[ \begin{matrix} -483&2618&164&-1&581&-2389\\
       1292&-1668&3245&-2392&-799&2144\\
       1867&-6848&-4082&2232&-8988&7034\\
       -281&121&-86&163&151&-262\\
       135&300&-280&262&230&-112\\
       585&-451&-395&260&107&-328
        \end{matrix} \right] }\\
        \end{matrix}
\end{equation}
\begin{figure*}[!t]
\centering
\includegraphics[width=1.6\columnwidth]{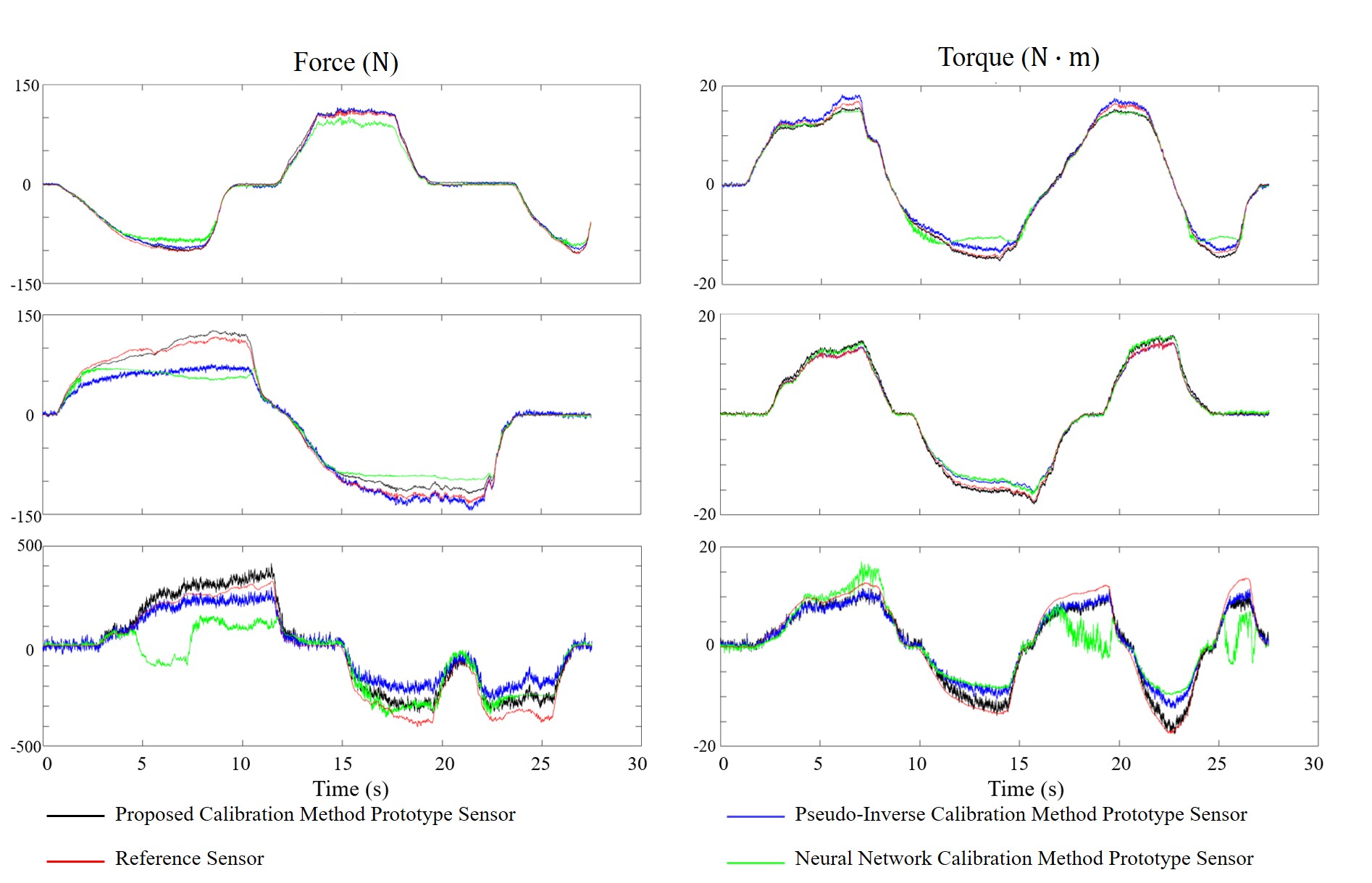}
\caption{Comparison Results of Calibration Methods in Terms of Time Domain Response: The black line represents the values measured using the proposed calibration method, while the red line indicates the reference sensor, the blue line illustrates the pseudo-inverse method, and the green line shows the neural network. These measurements were taken simultaneously under the same applied forces, and it can be observed that the proposed calibration method exhibits the least cross-talk.
}
\label{fig11}
\end{figure*}

\begin{equation}
    \begin{matrix}
       \ A_{proposed} =\Scale[0.8]{\left[ \begin{matrix}
           885&-187&884&242&-1774&-0\\
           2398&148&-2362&148&41&-231\\
           -2418&-2329&-3718&-161&-2156&-367\\
           -308&-78&355&-85&220&-232\\
           173&77&83&103&-274&-2\\
           444&-810&438&-310&442&-530
       \end{matrix} \right]}

    \end{matrix}
\end{equation}
In these equations, $A_{pse}$ (9) refers to the calibration matrix derived using pseudo inverse, and $A_{proposed}$ (10) refers to the calibration matrix computed using the proposed calibration method. Upon performing the calibration, a calibration matrix is generated. The result attained through pseudo inversion differs from that derived using Finite Element Method (FEM), mainly because the null space isn't accounted for in the former. Consequently, there is a marked difference between the outcomes obtained via pseudo inverse and those derived using the proposed method.

As shown in Equation (9), for example, the z-axis force calibration matrix's values should possess the same sign. However, $A_{\mathit{pse}}$'s z-axis force calibration matrix values are $\left[\begin{matrix}\begin{matrix}1867&-6848\\\end{matrix}&\begin{matrix}-4082&2232\\\end{matrix}&\begin{matrix}-8988&7034\\\end{matrix}\\\end{matrix}\right]$.

This signifies that the null space of the z-axis affects the result of the calibration matrix.
As a comparison group, a neural network and the ensuing results were also obtained. A feedforward net was used with one input layer (6 inputs), one output layer (6 outputs), and one hidden layer (12 nodes). Calibration was performed on the same data for 30 seconds per axis, and the results were compared as follows.

\subsection{Calibration Comparison}
\begin{table*}[!t]
\begin{center}
\caption{Comparison With Percentage Error and Non-Linearity(FSE)}
\label{tab9}
\resizebox{2\columnwidth}{!}{
\begin{tabular}{c c c c c c c c c c c c c}
\hline\hline
  &\multicolumn{4}{c}{\bf{Proposed Method}} & \multicolumn{4}{c}{Pseudo-Inverse} & \multicolumn{4}{c}{Neural Network} \\
  &Mean &Std &Max&Non-Linearity& Mean &Std &Max&Non-Linearity&Mean &Std &Max&Non-Linearity  \\ \hline
  \ $ F_x $ &\bf{0.13}&0.068&0.36&\bf{0.58}& 0.13&0.082&0.41&0.66&0.27&0.28&0.96&1.89   \\
  \ $ F_y $ &\bf{0.26}&0.19&0.67&\bf{1.12}&       0.58&0.62&2.12&2.72&            0.81&0.75&2.71&4.21  \\
  \ $ F_z $ &\bf{0.68}&0.46&2.37&\bf{3.15}&      1.03&0.96&3.81&5.69&            1.37&1.49&6.11 &8.43  \\
  \ $ M_x $ &\bf{0.72}&0.60&2.72&\bf{4.20}&           0.89& 0.67&3.39&4.80        &1.43   &1.62    &7.21&10.75\\
  \ $ M_y $ &\bf{0.92}& 0.70&3.35&\bf{5.48}&      0.63&0.77&3.48&4.53&     1.04&0.95&3.90&5.43    \\
  \ $ M_z $ &\bf{1.32}& 0.95&6.01&\bf{9.27}&      1.82&1.36&6.59&10.97&     2.88&3.18&14.46&22.00   \\
  \hline
\end{tabular}}
\end{center}
\end{table*}
The proposed method was compared to the most commonly used pseudo inverse matrix method and the neural network method. The results are shown in terms of the time-domain responses of each axis and the percent errors for accuracy and nonlinearity.
The time-domain results of the comparison among methods are shown in Fig.~\ref{fig11}.
The percent errors are calculated in terms of full-scale error as TABLE~\ref{tab9}. The maximum improvement of our proposed method is approximately 2 times that of the pseudoinverse and neural network. As TABLE~\ref{tab9} shows, the nonlinearity of the proposed calibration method did not exceed 10\%. However, the pseudo-inverse and neural networks present at most twice as much nonlinearity as the proposed method. This indicates that the proposed calibration method can guarantee performance even in nonlinear response conditions. Note that FSE represents the full-scale error.

\begin{table*}[!t]
\begin{center}
\caption{Comparison With other F/T Sensors for robots}
\label{tab11}
\resizebox{\linewidth}{!}{
\begin{tabular}{c c c c c c c c c c}
\hline\hline
  & Accuracy(FSE(\%)) &Sampling Rate&Resolution w/o external DAQ&Cost&Contact\\

 \hline
 \cellcolor[HTML]{EFEFEF} \bf{Proposed Method } &\cellcolor[HTML]{EFEFEF}2.0 &\cellcolor[HTML]{EFEFEF}5,000Hz&\cellcolor[HTML]{EFEFEF}14-bit&\cellcolor[HTML]{EFEFEF}Low&\cellcolor[HTML]{EFEFEF}Noncontact\\

Capacitor type~\cite{ref13}&4.1&200Hz&11-bit&Low&Noncontact\\
Magnetic type~\cite{ref14}&22.5&2ms delay&Low&Handmade&Noncontact\\
Pressure Sensing type~\cite{ref15}&22.2&1kHz&Low&Handmade&Contact\\
  RFT40-SA01(Commercial) &3.0  &200Hz&11-bit&Low-Medium&Noncontact\\
  SensOne-Serial(Commercial) &2.0  &800Hz&11-bit&Medium&Contact\\
    \hline
   & Weight&Load Capacity &Number of Sensors &Number of Parts\\
  \hline
 \cellcolor[HTML]{EFEFEF} \bf{Proposed Method } &\cellcolor[HTML]{EFEFEF}79g&\cellcolor[HTML]{EFEFEF}3700N &\cellcolor[HTML]{EFEFEF}6 &\cellcolor[HTML]{EFEFEF}4 with ADC \\

Capacitor type~\cite{ref13}&15g&100N&6&7 with ADC\\
Magnetic type~\cite{ref14}&-&1000N&5(2DOF)&4 without ADC\\
Pressure Sensing type~\cite{ref15}&-&450N&9(3DOF)&2 without ADC\\
  RFT40-SA01(Commercial) &60g&300N&6&- \\
  SensOne-Serial(Commercial)& 235g&2400N&12~&4 \\

  \hline
\end{tabular}}
\end{center}
\end{table*}
\section{Discussion}

Two experiments were performed: one for determining the specification and performance requirements of the proposed sensor; the other for comparing the proposed calibration method's performance against existing calibration methods.
A first observation is that the specifications of the prototype sensor are several percent better in terms of accuracy when compared with other studies. When compared with commercial sensors, our proposed method's differences and advantages include a fast response of 5 kHz and a high resolution as TABLE~\ref{tab11} .
In addition, the number of parts of our proposed sensor is small, making the cost low. The total cost with 4 parts is below \$250.

\begin{figure}[!t]
\centering
\includegraphics[width=0.7\columnwidth]{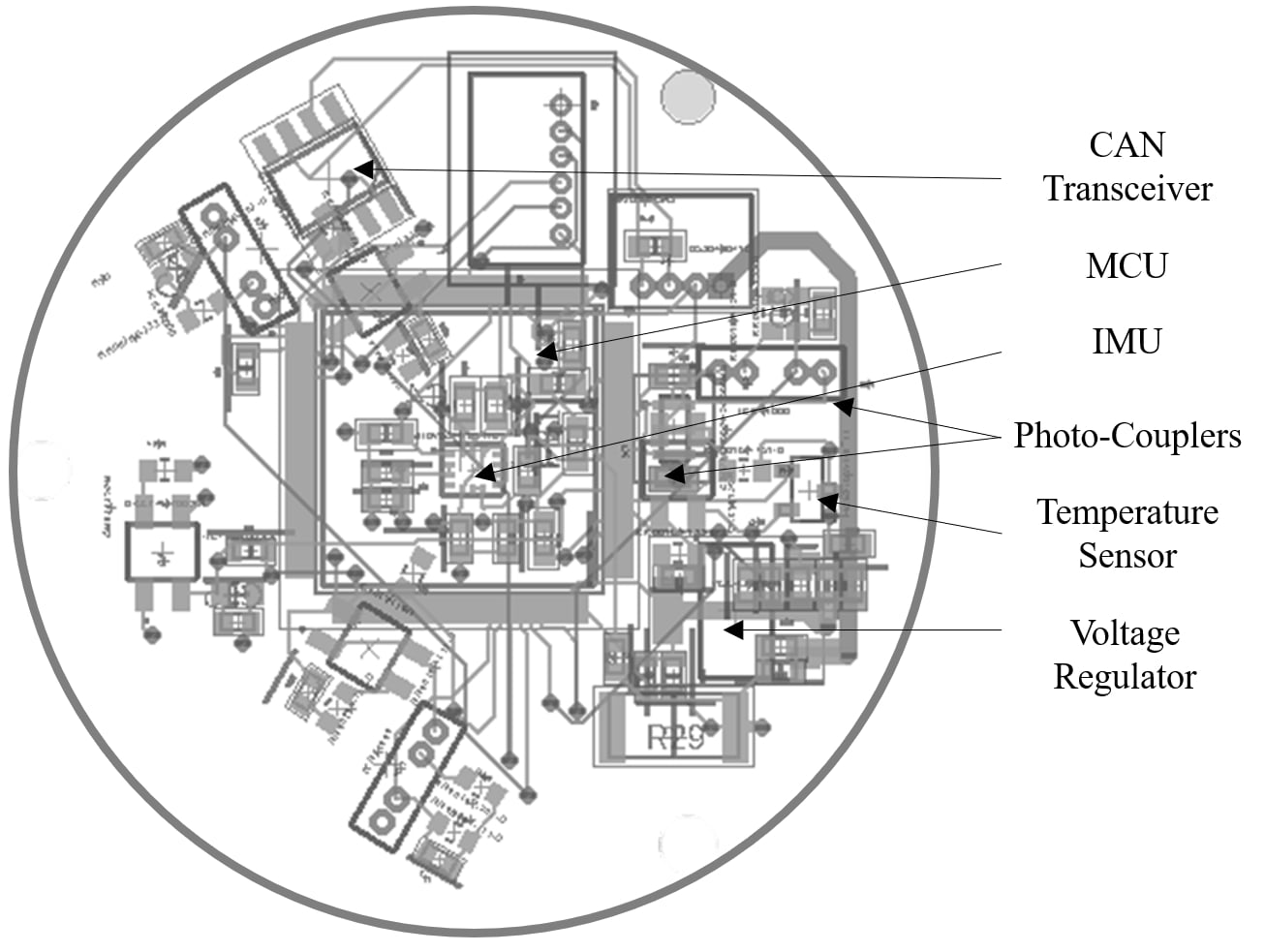}
\caption{Printed Circuit Board with MCU, IMU, and CAN Transceiver: The PCB is furnished with an IMU, temperature sensor, MCU, and CAN transceiver, enabling direct communication and providing the capability for future program updates. Additionally, the ample space available on the PCB allows for potential further miniaturization.
}
\label{fig12}
\end{figure}

Moreover, since the size of the sensor is small, it is possible to include an MCU (micro controller unit), communication chip, and even an IMU (inertia measurement unit), as shown in Fig.~\ref{fig12}. Since only 6 sensors are included,  the cost is less than that of a system using 12 sensors, with the cost of the photocoupler itself also being low. In addition, the full device weighs less than 80 g and is noncontact, leaving the sensor strong against impact. Additionally, since there is no labor involved in making the sensor, the cost of the relative sensor will further reduce in the case of mass production. However, there is exists a nonlinearity of the sensor itself and differences in the gain value of the photocoupler due to resistance errors.

Also, the proposed sensor's performance will be improved by using the Darlington transistor, which enables easier and faster operation with improved accuracy due to the better amplification than when using the normal transistor method. 

Given that the structure consists of a cross elastic beam, its nonlinearity is more pronounced than in other structures. Using designs with less nonlinearity could potentially enhance performance. While AL7075-T6 material was employed in this study, the inherent nonlinearity might still persist due to its non-steel nature. Thus, it is anticipated that the nonlinearity could be reduced by utilizing steel materials such as AISI4340 steel, instead of AL7075-T6. The cross elastic beam structure was chosen largely due to the prevalence of sensors employing this design, and to showcase the advantages of optical sensors over the strain gauge method. By employing a structure that allows for the decoupling of each force and torque axis, the application process is simplified and cross-talk is reduced.

During calibration, the experiment was conducted as fast as 30 seconds. 
The proposed method demonstrated the highest degree of accuracy and mitigated nonlinearity. Furthermore, this method can address the issue of null spaces prevalent in multi-axis F/T sensors. The results indicate that the optimization problem-solving approach performs well, even when applied to system identification. This method suggests that calibration can be achieved cost-effectively and efficiently when the structure is well-defined and there is no need for additional equipment.

\section{Conclusion}

In this paper, a novel 6-axis F/T sensor principle was presented using a photocoupler. The 6-axis F/T sensor uses the reflection-type photocoupler and has a form that can measure the vertical and horizontal forces by using the spring as a screen. Thus, the sensor records 6-axis force and torque. It is designed on one PCB and to be made on one plane. It shows good accuracy in the 6-axis F/T implementation, with a fast 5 kHz bandwidth. The proposed sensor can be easily assembled by bolting of only three mechanical plate-type parts and one printed circuit board; total 4 parts including PCB. The assembly process is simple and mass production and customization are possible. The cost of the sensor is below \$250; compared to a comparable commercial sensor, it has a cost of 1/20. Because it uses a noncontact method, it further has strong durability and light weight. Because it uses a small sensor, it is capable of miniaturization and further cost efficiency. Since the signal is amplified by the sensing element, it does not require a separate amplifier, making it is possible to transmit signals by inserting an IMU, MCU, or CAN (Controller Area Network) transceiver. In addition, a method for system identification using an optimization solver and a calibration method using a known structure is presented. Accurate calibration can be performed quickly with low cost and effort and without expensive additional equipment. Due to its novelty and utility, this sensor has been disclosed in a patent application. Furthermore, this sensor and calibration method are expected to be used in many instances, particularly where robotic applications are popular.

\section*{APPENDIX}
\begin{figure}[!b]
\centering
\includegraphics[width=\columnwidth]{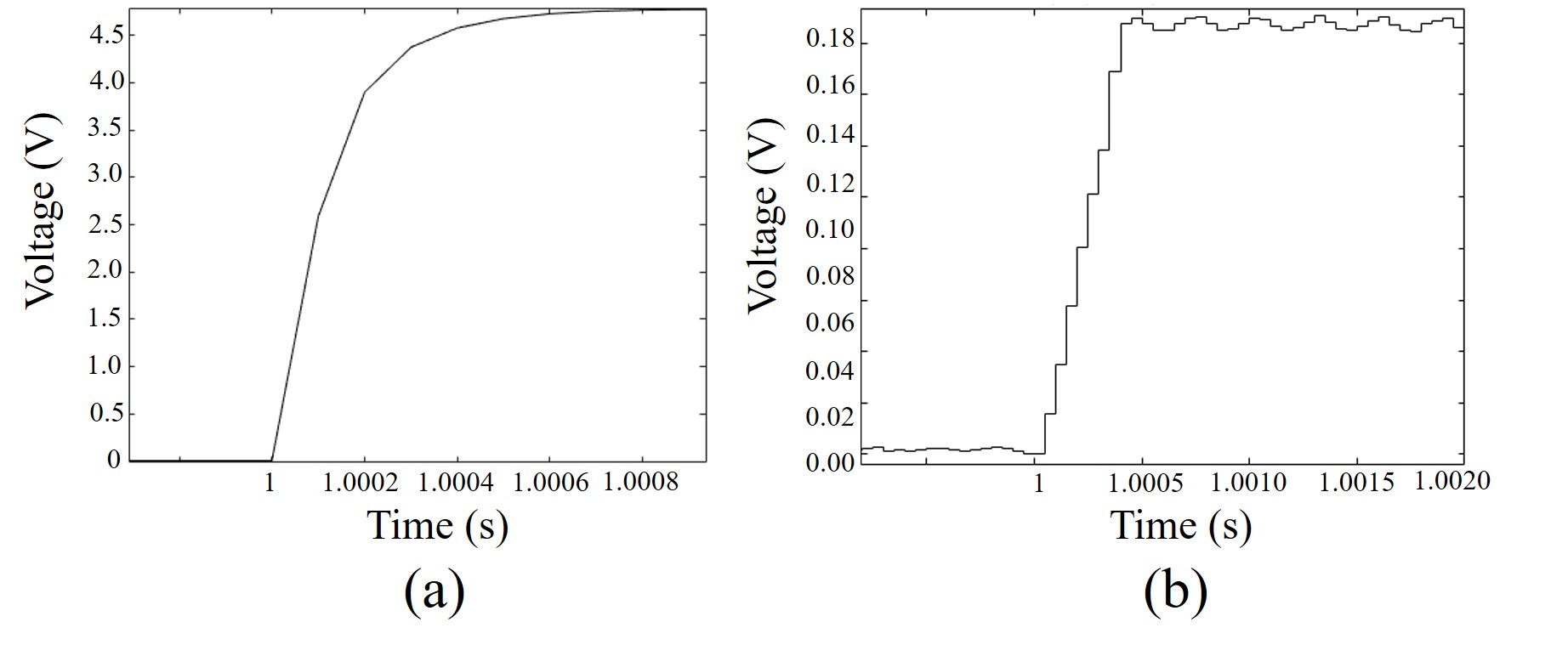}
\caption{This shows each sensor's rise time that measured by using step input of the LED of each sensor. The rise time of the sensor can be adjusted by changing the value of the resistor which is connected with photo-transistor.\\(a) Rise time of the QRE1113GR (b) Rise time of the TCRT1000.}
\label{fig7}
\end{figure}

TABLE~\ref{tab2} presents the parameters of the cross-elastic beam in Section \uppercase\expandafter{\romannumeral2}, Sensor Design and Fabrication. The number of each alphabet refers Fig.~\ref{fig5}.
The sensors' rise time experiments were done using step input and the graph was shown as Fig.~\ref{fig7}. From this results, rise time of each sensor can be calculated in Section \uppercase\expandafter{\romannumeral3}.A. 

\begin{table}[!b]
\begin{center}
\caption{Value of the Cross-Elastic Beam's Parameter}
\label{tab2}
\begin{tabular}{c c c c c c c}
\hline\hline
 parameter    & $l_1$     & $l_2$     & $b_1$     & $b_2$     & $h_1$     & $h_2$     \\ \hline
value(mm) & 12 & 21 & 4 & 0.5 & 6 & 6 \\
 \hline
\end{tabular}
\end{center}
\end{table}

\begin{figure}[!b]
\centering
\includegraphics[width=\columnwidth]{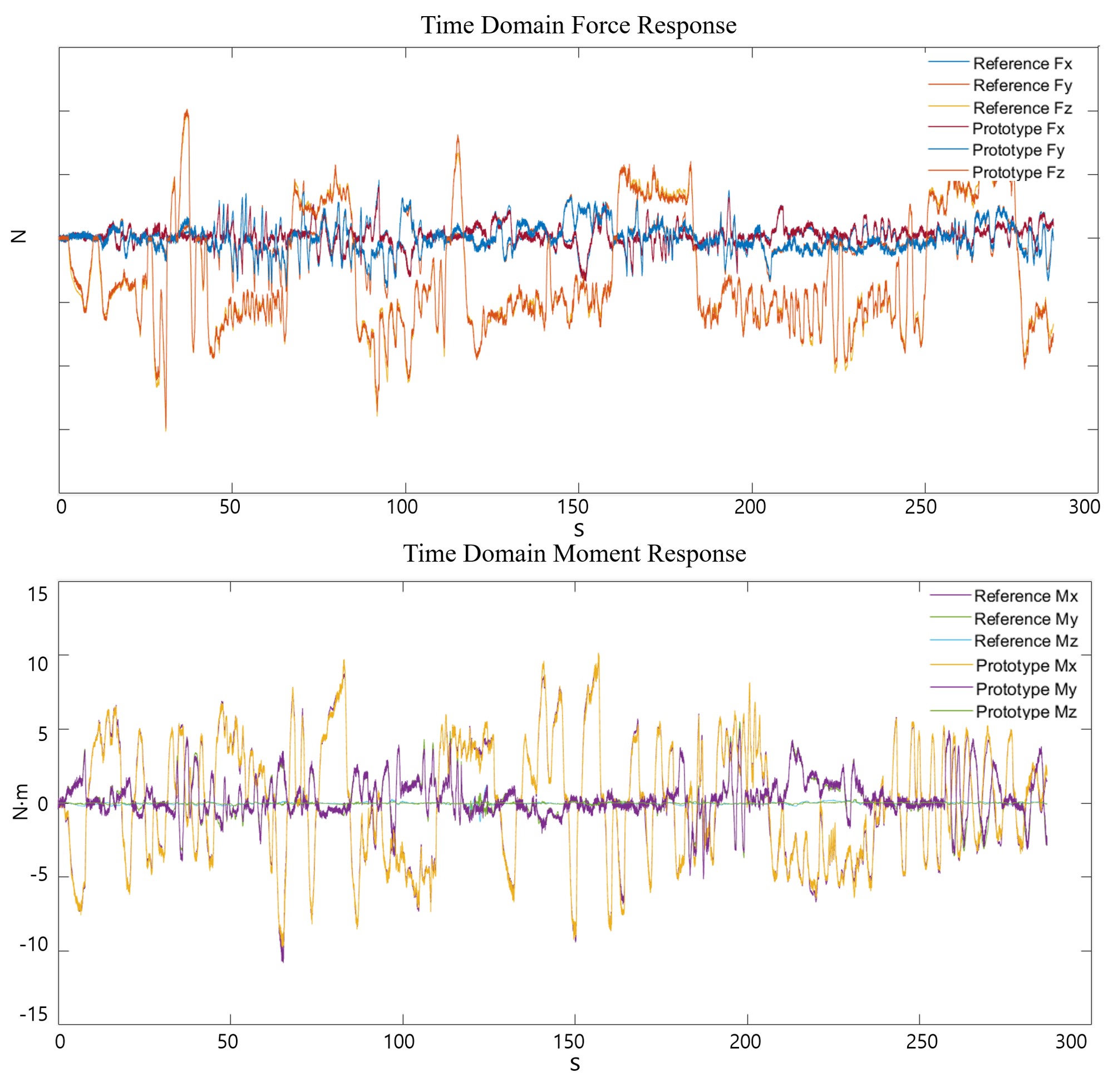}
\caption{Comparison result in terms of time domain response between prototype and reference sensors. The forces and moments are applied at the same time.}
\label{fig20}
\end{figure}
An experiment was also carried out, involving multiple axes simultaneously. The outcome, as depicted in Fig.\ref{fig20}, indicates positive results. The rise time experiments for the sensors were conducted using a step input, with the results illustrated in Fig.\ref{fig7}. It shows the low cross-talk which was referred in Section \uppercase\expandafter{\romannumeral4}.A.


\end{document}